\documentclass[11pt]{article}
\usepackage{acl2016}
\usepackage{times}
\usepackage{latexsym}
\usepackage{url}
\usepackage{amsmath}
\usepackage{algorithm}
\usepackage{algpseudocode}

\aclfinalcopy 

\newcommand{\argmax}{\arg\!\max}

\title{Sense Embedding Learning for Word Sense Induction}

\author{Linfeng Song$^1$, Zhiguo Wang$^2$, Haitao Mi$^2$ \and Daniel Gildea$^1$ \\
  $^1$Department of Computer Science, University of Rochester, Rochester, NY 14627 \\
  $^2$IBM T.J. Watson Research Center, Yorktown Heights, NY 10598  \\
  }

\date{}

\begin{document}

\maketitle

\begin{abstract}
Conventional word sense induction (WSI) methods usually represent each instance with discrete linguistic features or co-occurrence features, and train a model for each polysemous word individually. 
In this work, we propose to learn sense embeddings for the WSI task. 
In the training stage, our method induces several sense centroids (embedding) for each polysemous word. 
In the testing stage, our method represents each instance as a contextual vector, and induces its sense by finding the nearest sense centroid in the embedding space. 
The advantages of our method are 
(1) distributed sense vectors are taken as the knowledge representations which are trained discriminatively, and usually have better performance than traditional count-based distributional models,
and (2) a general model for the whole vocabulary is jointly trained to induce sense centroids under the mutli-task learning framework. 
Evaluated on SemEval-2010 WSI dataset, our method outperforms all participants and most of the recent state-of-the-art methods.
We further verify the two advantages by comparing with carefully designed baselines.
\end{abstract}

\section{Introduction}

Word sense induction (WSI) is the task of automatically finding sense clusters for polysemous words. In contrast, 
word sense disambiguation (WSD) assumes there exists an already-known sense inventory, 
and the sense of a word type is disambiguated according to the sense inventory.
Therefore, clustering methods are generally applied in WSI tasks, while classification methods are utilized in WSD tasks.
WSI has been successfully applied to many NLP tasks such as machine translation \cite{xiong-zhang:2014:P14-1}, information retrieval \cite{navigli-crisafulli:2010:EMNLP} and novel sense detection \cite{lau-EtAl:2012:EACL2012}.

However, existing methods usually represent each instance with discrete hand-crafted features \cite{bordag2006word,chen-EtAl:2009:NAACLHLT091,vandecruys-apidianaki:2011:ACL-HLT2011,purandare-pedersen:2004:CONLL}, 
which are designed manually and require linguistic knowledge.
Most previous methods require learning a specific model for each polysemous word, which limits their usability for down-stream applications and loses the chance to jointly learn senses for multiple words.

There is a great advance in recent distributed semantics, such as word embedding \cite{mikolov2013distributed,pennington-socher-manning:2014:EMNLP2014} and 
sense embedding \cite{reisinger-mooney:2010:NAACLHLT,huang-EtAl:2012:ACL20122,jauhar-dyer-hovy:2015:NAACL-HLT,rothe-schutze:2015:ACL-IJCNLP,chen-liu-sun:2014:EMNLP2014,tian-EtAl:2014:Coling}.
Comparing with word embedding, sense embedding methods learn distributed representations for senses of a polysemous word, which is similar to the sense centroid of WSI tasks. 

In this work, we point out that the WSI task and the sense embedding task are highly inter-related, and propose to jointly learn sense centroids (embeddings) of all polysemous words for the WSI task. 
Concretely, our method induces several sense centroids (embedding) for each polysemous word in training stage. 
In testing stage, our method represents each instance as a contextual vector, and induces its sense by finding the nearest sense centroid in the embedding space.
Comparing with existing methods, our method has two advantages: 
(1) distributed sense embeddings are taken as the knowledge representations which are trained discriminatively, and usually have better performance than traditional count-based distributional models \cite{baroni-dinu-kruszewski:2014:P14-1},
and (2) a general model for the whole vocabulary is jointly trained to induce sense centroids under the mutli-task learning framework \cite{caruana1997multitask}.
Evaluated on SemEval-2010 WSI dataset, our method outperforms all participants and most of the recent state-of-the-art methods.

\section{Methodology}


\subsection{Word Sense Induction}

WSI is generally considered as an unsupervised clustering task under the distributional hypothesis \cite{harris1954distributional} that the word meaning is reflected by the set of contexts in which it appears. 
Existing WSI methods can be roughly divided into feature-based or Bayesian.
Feature-based methods first represent each instance as a context vector, then utilize a clustering algorithm on the context vectors to induce all the senses.
Bayesian methods \cite{brody-lapata:2009:EACL,yao2011nonparametric,lau-EtAl:2012:EACL2012,goyal2014unsupervised,TACL485}, on the other hand, discover senses based on topic models.
They adopt either the LDA \cite{LDA:2003} or HDP \cite{HDP:2006} model by viewing each target word as a corpus and the contexts as pseudo-documents, where a context includes all words within a window centred by the target word.
For sense induction, they first extract pseudo-documents for the target word, then train topic model, finally pick the most probable topic for each test pseudo-document as the sense.

All of the existing WSI methods have two important factors: 1) how to group similar instances (clustering algorithm) and 2) how to represent context (knowledge representation).
For clustering algorithms, 
feature-based methods use k-means or graph-based clustering algorithms to assign each instance to its nearest sense, 
whereas Bayesian methods sample the sense from the probability distribution among all the senses for each instance, 
which can be seen as soft clustering algorithms. 
As for knowledge representation, existing WSI methods use the vector space model (VSM) to represent each context. 
In feature-based models, each instance is represented as a vector of values, where a value can be the count of a feature or the co-occurrence between two words.
In Bayesian methods, 
the vectors are represented as co-occurrences between documents and senses or between senses and words.
Overall existing methods separately train a specific VSM for each word.
No methods have shown distributional vectors can keep knowledge for multiple words while showing competitive performance.

\subsection{Sense Embedding for WSI}


As mentioned in Section 1, sense embedding methods learn a distributed representation for
each sense of a polysemous word. There are two key factors for sense embedding learning: (1) how to decide the number of senses for each polysemous word and (2) how to learn an embedding representation for each sense. 
To decide the number of senses in factor (1), one group of methods \cite{huang-EtAl:2012:ACL20122,neelakantan-EtAl:2014:EMNLP2014} set a fixed number \emph{K} of senses for each word, and each instance is assigned to the most probable sense according to Equation \ref{eq:kmeans}, where $\mu(w_t,k)$ is the vector for the $k$-th sense centroid of word $w$, and $v_c$ is the representation vector of the instance.

\begin{equation} \label{eq:kmeans}
    s_t = \argmax_{k=1,..,K} sim(\mu(w_t,k),v_c)
\end{equation}

Another group of methods \cite{li-jurafsky:2015:EMNLP} employs non-parametric algorithms to dynamically decide the number of senses for each word, and 
each instance is assigned to a sense following a probability distribution in Equation \ref{eq:nonp}, where $S_t$ is the set of already generated senses for $w_t$, and $\gamma$ is a constant probability for generating a new sense for $w_t$.

\begin{equation} \label{eq:nonp}
s_t  \sim \begin{cases}
               p(k|\mu(w_t,k),v_c)~\forall~k \in S_t \\
               \gamma~~\text{for new sense}
            \end{cases}
\end{equation}

From the above discussions, we can obviously notice that WSI task and sense embedding task are inter-related. The two factors in sense embedding learning can be aligned to the two factors of WSI task. Concretely, deciding the number of senses is the same problem as the clustering problem in WSI task, and sense embedding is a potential knowledge representation for WSI task.
Therefore, sense embedding methods are naturally applicable to WSI.

In this work, we apply the sense embedding learning methods for WSI tasks. Algorithm \ref{algo:online} lists the flow of our method. The algorithm iterates several times over a Corpus (Line 2-3).
For each token $w_t$, 
it calculates the context vector $v_c$ (Line 4) for an instance, 
and then gets the most possible sense label $s_t$ for $w_t$ (Line 5). 
Finally, both the sense embeddings for $s_t$ and global word embeddings for all context words of $w_t$ are updated (Line 6).
We introduce our strategy for \emph{context\_vec} in the next section.
For \emph{sense\_label} function, a sense label is obtained by either Equation \ref{eq:kmeans} or Equation \ref{eq:nonp}.
For the \emph{update} function, vectors are updated by the Skip-gram method (same as \newcite{neelakantan-EtAl:2014:EMNLP2014}) which tries to predict context words with the current sense. 
In this algorithm, the senses of all polysemous words are learned jointly on the whole corpus, instead of training a single model for each individual word as in the traditional WSI methods. 
This is actually an instance of multi-task learning, where WSI models for each target word are trained together, and all of these models share the same global word embeddings.

\begin{algorithm}[t]
  \caption{\small{Sense Embedding Learning for WSI}}\label{algo:online}
  \begin{algorithmic}[1]  
    \Procedure{Training}{Corpus $C$}
    \For{\texttt{$iter$ in [$1..I$]}}
      \For{\texttt{$w_t$ in $C$}}
        \State\texttt{$v_c$ $\leftarrow$ context\_vec($w_t$)}
        \State\texttt{$s_t$ $\leftarrow$ sense\_label($w_t$, $v_c$)}
        \State\texttt{update($w_t$, $s_t$)}
      \EndFor
    \EndFor
    \EndProcedure
  \end{algorithmic}
\end{algorithm}

Comparing to the traditional methods for WSI tasks, the advantages of our method include: 
1) WSI models for all the polysemous words are trained jointly under the multi-task learning framework;
2) distributed sense embeddings are taken as the knowledge representations which are trained discriminatively, and usually have better performance than traditional count-based distributional models \cite{baroni-dinu-kruszewski:2014:P14-1}.
To verify the two statements, we carefully designed comparative experiments described in the next section.

\section{Experiment}

\subsection{Experimental Setup and baselines}

\begin{table*}
    \centering
    \begin{tabular}{|l|l|l|l|l|l|l|l|l|l|l|l|} 
       \hline
       System     & \multicolumn{3}{c|}{V-Measure(\%)} & \multicolumn{3}{c|}{Paired F-score(\%)} & \multicolumn{3}{c|}{80-20 SR(\%)} & FS &\#CI  \\
       \hline
                  & All  & Noun & Verb & All  & Noun & Verb & All  & Noun & Verb & All & \\
       \hline
        UoY~(2010) & 15.7 & 20.6 & 8.5  & 49.8 & 38.2 & 66.6 & 62.4 & 59.4 & 66.8 & -    & 11.5 \\
NMF$_{lib}$~(2011) & 11.8 & 13.5 & 9.4  & 45.3 & 42.2 & 49.8 & 62.6 & 57.3 & 70.2 & -    & 4.80  \\
         NB~(2013) & \textbf{18.0} & \textbf{23.7} & \textbf{9.9}  & 52.9 & 52.5 & 53.5 & 65.4 & 62.6 & 69.5 & -    & 3.42  \\
   Spectral~(2014) & 4.5  & 4.6  & 4.2  & \textbf{61.5} & \textbf{54.5} & \textbf{71.6} & -    & -    & -    & 60.7 & 1.87  \\
       \hline
   SE-WSI-fix-cmp  & 16.3  & 20.8 & 9.7  & 54.3 & 54.2 & 54.3 & \textbf{66.3} & \textbf{63.6} & \textbf{70.2}  & \textbf{66.4} & 2.61  \\
    SE-WSI-fix    & 9.8  & 13.5 & 4.3  & 55.1 & 50.7 & 61.6 & 62.9 & 58.5 & 69.2  & 63.0 & 2.50  \\
    SE-WSI-CRP    & 5.7  & 7.4  & 3.2  & 55.3 & 49.4 & 63.8 & 61.2 & 56.3 & 67.9  & 61.3 & 2.09  \\
       CRP-PPMI   & 2.9  & 3.5  & 2.0  & 57.7 & 53.3 & 64.0 & 59.2 & 53.6 & 67.4  & 59.2 & 1.76  \\
       WE-Kmeans  & 4.6  & 5.0  & 4.1  & 51.2 & 46.5 & 57.6 & 58.6 & 53.3 & 66.4  & 58.6 & 2.54  \\
       \hline
    \end{tabular}
    \caption{{Result on SemEval-2010 WSI task. \emph{80-20 SR} is the supervised recall of 80-20 split supervised evaluation. \emph{FS} is the F-Score of 80-20 split supervised evaluation. \emph{\#CI} is the average number of clusters (senses) }}
    \label{tab:semeval_2010}
\end{table*}

We evaluate our methods on the test set of the SemEval-2010 WSI task \cite{manandhar2010semeval}.
It contains 8,915 instances for 100 target words (50 nouns and 50 verbs) which mostly come from news domain.
We choose the April 2010 snapshot of Wikipedia \cite{wiki2010} as our training set, as it is freely available and domain general.
It contains around 2 million documents and 990 million tokens.
We train and test our models and the baselines according to the above data setting, and compare with reported performance on the same test set from previous papers.

For our sense embedding method, we build two systems: 
\emph{SE-WSI-fix} which adopts Multi-Sense Skip-gram (MSSG) model \cite{neelakantan-EtAl:2014:EMNLP2014} and assigns 3 senses for each word type,
and \emph{SE-WSI-CRP} \cite{li-jurafsky:2015:EMNLP} which dynamically decides the number of senses using a Chinese restaurant process.
For \emph{SE-WSI-fix}, we learn sense embeddings for the top 6K frequent words in the training set. 
For \emph{SE-WSI-CRP}, we first learn word embeddings with word2vec\footnote{\texttt{https://code.google.com/p/word2vec/}}, 
then use them as pre-trained vectors to learn sense embeddings.
All training is under default parameter settings, and all word and sense embeddings are fixed at 300 dimensions.
For fair comparison, we create \emph{SE-WSI-fix-cmp} by training the MSSG model on the training data of the SemEval-2010 WSI task with the same setting of \emph{SE-WSI-fix}.

We also design baselines to verify the two advantages of our sense embedding methods.
One (\emph{CRP-PPMI}) uses the same CRP algorithm as \emph{SE-WSI-CRP}, but with Positive PMI vectors as pre-trained vectors.
The other (\emph{WE-Kmeans}) uses the vectors learned by \emph{SE-WSI-fix}, but separately clusters all the context vectors into 3 groups for each target word with kmeans.
We compute a context vector by averaging the vectors of all selected words in the context\footnote{A word is selected only if its length is greater than 3, not the target word, or not in a self-constructed stoplist.}.

\subsection{Comparing on SemEval-2010}

We compare our methods with the following systems: (1) 
\emph{UoY} \cite{korkontzelos-manandhar:2010:SemEval} which is the best system in the SemEval-2010 WSI competition;
(2) \emph{NMF$_{lib}$} \cite{vandecruys-apidianaki:2011:ACL-HLT2011} which adopts non-negative matrix factorization to factor a matrix and then conducts word sense clustering on the test set; 
(3) \emph{NB} \cite{choe-charniak:2013:EMNLP} which adopts naive Bayes with the generative story that a context is generated by picking a sense and then all context words given the sense; 
and (4) \emph{Spectral} \cite{goyal2014unsupervised} which applies spectral clustering on a set of distributional context vectors.

Experimental results are shown in Table \ref{tab:semeval_2010}. Let us see the results on supervised recall (80-20 SR) first, as it is the main indicator for the task. 
Overall, \emph{SE-WSI-fix-cmp}, which jointly learns sense embedding for 6K words, outperforms every comparing systems which learns for each single word. 
This shows that sense embedding is suitable and promising for the task of word sense induction.
Trained on out-of-domain data, \emph{SE-WSI-fix} outperforms most of the systems, including the best system in the shared task (\emph{UoY}), and
\emph{SE-WSI-CRP} works better than \emph{Spectral} and all the baselines.
This also shows the effectiveness of the sense embedding methods.
Besides, \emph{SE-WSI-CRP} is 1.7 points lower than \emph{SE-WSI-fix}.
We think the reason is that \emph{SE-WSI-CRP} induces fewer senses than \emph{SE-WSI-fix} (see the last column of Table \ref{tab:semeval_2010}).
Since both systems induce fewer senses than the golden standard which is 3.85, inducing fewer senses harms the performance.
Finally, simple as it is, \emph{NB} shows a very good performance.
However \emph{NB} can not benefit from large-scale data as its number of parameters is small,
and it uses EM algorithm which is generally slow. 
Sense embedding methods have other advantages that they train a general model while \emph{NB} learns specific model for each target word.

As for the unsupervised evaluations, \emph{SE-WSI-fix} achieves a good V-Measure score (VM) with a few induced senses. 
\newcite{pedersen:2010:SemEval} points out that bad models can increase VM by increasing the number of clusters, but doing this will harm performance on both Paired F-score (PF) and SR.
Even though \emph{UoY}, \emph{NMF$_{lib}$} and \emph{NB} show better VM, they (especially \emph{UoY}) induced more senses than \emph{SE-WSI-fix}.
\emph{SE-WSI-fix} has higher PF than all others, and higher SR than \emph{UoY} and \emph{NMF$_{lib}$}.
Trained on the official training data of SemEval-2010 WSI task, \emph{SE-WSI-fix-cmp} achieves the top performance on both VM and PF, while it induces a reasonable number of averaged senses.
Comparatively \emph{SE-WSI-CRP} has lower VM and induces fewer senses than \emph{SE-WSI-fix}.
One possible reason is that the ``rich gets richer'' nature of CRP makes it conservative for making new senses.
But its PF and SR show that it is still a highly competitive system.

To verify the advantages of our method, we first compare \emph{SE-WSI-CRP} with \emph{CRP-PPMI} as their only difference is the vectors for representing contexts.
We can see that \emph{SE-WSI-CRP} performs significantly better than \emph{CRP-PPMI} on both SR and VM.
\emph{CRP-PPMI} has higher PF mainly because it induces fewer number of senses.
The above results prove that using sense embeddings have better performance than using count-based distributional models.
Besides, \emph{SE-WSI-fix} is significantly better than \emph{WE-Kmeans} on every metric.
As \emph{WE-Kmeans} and \emph{SE-WSI-fix} learn sense centroids in the same vectors space, while the latter performs joint learning.
Therefore, the joint learning is better than learning separately.

\section{Related Work}

\newcite{krageback-EtAl:2015:VSM-NLP} proposed two methods to utilize distributed representations for the WSI task.
The first method learned centroid vectors by clustering all pre-computed context vectors of each target word.
The other method simply adopted \emph{MSSG} \cite{neelakantan-EtAl:2014:EMNLP2014} and changed context vector calculation from the average of all context word vectors to weighted average.
Our work has further contributions. 
First, we clearly point out the two advantages of sense embedding methods: 
1) joint learning under the mutli-task learning framework, 
2) better knowledge representation by discriminative training, and 
verify them by experiments.
In addition, we adopt various sense embedding methods to show that sense embedding methods are generally promising for WSI, not just one method is better than other methods.
Finally, we compare our methods with recent state-of-the-art WSI methods on both supervised and unsupervised metrics.

\section{Conclusion}

In this paper, we show that sense embedding is a promising approach for WSI by adopting two different sense embedding based systems on the SemEval-2010 WSI task.
Both systems show highly competitive performance while they learn a general model for thousands of words (not just the tested polysemous words).
we believe that the two advantages of our method are:
1) joint learning under the mutli-task learning framework, 
2) better knowledge representation by discriminative training, and 
verify them by experiments.

\section*{Acknowledgments}

Funded by NSF IIS-1446996. We would like to thank Yue Zhang for his insightful comments on the first version of the paper, and the anonymous reviewers for the insightful comments.

\bibliography{acl2016}
\bibliographystyle{acl2016}

\end{document}